\documentclass[runningheads]{llncs}

\usepackage{fancybox,fancyhdr,enumerate}
\usepackage[utf8]{inputenc}
\usepackage{graphicx}

\usepackage[perpage]{footmisc}

\usepackage{dsfont}
\usepackage{eurosym}
\usepackage{url}
\usepackage{color}
\usepackage{amsmath}
\usepackage{amsfonts}

\usepackage{gensymb} 

\usepackage[%
pdftitle={CIL-Technical Report},
pdfauthor={Tom Hanika},
pdfsubject={Collaborative Interactive Learning},
pdfcreator={MiKTeX, LaTeX with hyperref and KOMA-Script},
pdfkeywords={CIL},
pdfpagemode=UseOutlines,
pdfdisplaydoctitle=true,
pdflang=en,
pdfborder={0 0 0},
]{hyperref}

\usepackage{paralist}
\newenvironment{shortitemize}{\begin{compactitem}}{\end{compactitem}}

\begin{document}

\title{Collaborative Interactive Learning -- A clarification of terms and a differentiation from other research fields}

\author{Tom Hanika\inst{1} \and
        Marek Herde\inst{2} \and
        Jochen Kuhn\inst{3} \and
        Jan Marco Leimeister\inst{4} \and
		Paul Lukowicz\inst{5} \and
		Sarah Oeste-Rei{\ss}\inst{4} \and
		Albrecht Schmidt\inst{6} \and
		Bernhard Sick\inst{2} \and
        Gerd Stumme\inst{1} \and
		Sven Tomforde\inst{2}\and
	    Katharina Anna Zweig\inst{7}
}
		
\authorrunning{T. Hanika et al.}
\titlerunning{Collaborative Interactive Learning}

\institute{Knowledge \& Data Engineering, University of Kassel, Germany,\\
           \email{\{Tom.Hanika, stumme\}@cs.uni-kassel.de}
           \and
           Intelligent Embedded Systems, University of Kassel, Germany,\\
           \email{\{marek.herde, bsick, stomforde\}@uni-kassel.de}
           \and
           Didactics of Physics, Technical University of Kaiserslautern, Germany,\\
           \email{kuhn@physik.uni-kl.de}
           \and
           Information Systems, University of Kassel, Germany,\\
           \email{\{leimeister, oeste-reiss\}@uni-kassel.de}
           \and
           Embedded Intelligence, Technical University of Kaiserslautern, Germany,\\
           \email{Paul.Lukowicz@dfki.de}
           \and
           Human-Computer Interaction, Ludwig Maximilian University of Munich, Germany,\\
           \email{albrecht.schmidt@ifi.lmu.de}
           \and
           Algorithm Accountability, Technical University of Kaiserslautern, Germany,\\
           \email{zweig@cs.uni-kl.de}
}

\maketitle


\begin{abstract}
	The field of collaborative interactive learning (CIL) aims at developing and investigating the technological foundations for a new generation of smart systems that support humans in their everyday life. While the concept of CIL has already been carved out in detail (including the fields of dedicated CIL and opportunistic CIL) and many research objectives have been stated, there is still the need to clarify some terms such as information, knowledge, and experience in the context of CIL and to differentiate CIL from recent and ongoing research in related fields such as active learning, collaborative learning, and others. Both aspects are addressed in this paper.
\end{abstract}

\section{Introduction}

The ability to improvise, adapt to changing circumstances, and dynamically evolve new strategies in response to unexpected developments is a core human strength. While machine learning has made tremendous progress in the last decade, the ability of today's systems to  improvise and  adapt beyond what they have been trained to deal with at design time is still very limited.  Thus, for example, speech recognition systems   can improve beyond design time training by collecting more data and using it to adapt to a specific user.  However, this can hardly be considered adaptation to radically new circumstances. Instead, the systems rather fine-tune and optimize the functionality within a scope for which the systems were trained at design time. By contrast consider a car that was developed to drive autonomously in the orderly traffic conditions of a highway. Such a car is likely to fail when deployed within less structured, more collaborative conditions of city traffic. 

Key aspects of the human ability to adapt to new situations and tasks are the  preparedness for ``lifelong learning'' and the ability to leverage the experiences made by others through interactive collaboration with peers.
Thus, for example, most students coming from high school to university find that their acquired ``old''  studying strategies are inadequate. However, most are prepared to learn new ways and manage to quickly adapt to university life. To this end, they are heavily dependent on help and advice from higher semester fellow students, (which is often provided through, for example, mentoring programs) as well as on the experience of other ``beginners''.   When changing jobs people usually take time to adjust. Again helpful co-workers and interaction with other people facing a similar situation are essential for learning processes.

{\em How can such collaborative, interactive  lifelong learning strategies be adopted by autonomous digital systems?} This is the fundamental research question addressed by the new research area \textbf{Collaborative Interactive Learning (CIL)}. CIL aims at creating the foundation for new kinds of autonomous systems that can act in dynamic,  unconstrained real-world environments, are effective, understandable, and pleasant in their interaction and collaboration with humans.  CIL considers not just machine-machine collaboration  but also the questions how the learning process of digital systems can be  enhanced through seamless, unobtrusive interaction with humans and the other way round: how autonomous interactive systems  can enhance human collaborative learning. 
Towards the above vision, CIL conducts research into collaborative sensing and perception, algorithms and architectures, as well as interaction models on individual as well as societal level.  CIL will create systems that provide meaningful and desirable support for humans without taking autonomy from them or even dominating and ultimately replacing them.

Research on CIL  started in 2015 by carving out the various research challenges and solution statements in the field of dedicated CIL (D-CIL)~\cite{CLL+16}. 
Then, preliminary work in the field of opportunistic CIL (O-CIL) was evaluated, various application fields were discussed, and results of a case study on collaborative travel time estimation were presented~\cite{BCL+16}. After some further work on the CIL concept \cite{COS+18,SOS+17,CKL+18},
the authors prepared the work on CIL with joint work such as developing techniques for simulating human oracles~\cite{CS17}, analysis of the practical use of human oracles in active learning~\cite{CSK+18}, and performing a case study that evaluates how human oracles perform when labeling traffic signs~\cite{Cal17}. 
From a rather human centered point of view the authors prepared the work on CIL by developing a framework for a future reallocation of tasks between humans and machines~\cite{TOL17},
designing and evaluating collaborative work practices for enhancing human learning~\cite{Oes17,OBS17},
enriching the body of collaboration engineering methodologies by defining a research agenda when teammates are not human~\cite{SBB+18},
developing a new modeling technique~\cite{WBV+19},
and proposing a taxonomy of design option combinations for conversational agents in collaborative work~\cite{BOL19}.
The authors have also developed new knowledge acquisition methods~\cite{conf/icfca/BorchmannHO17,journals/corr/abs-1807-06149}, in particular in a collaborative setting~\cite{th17}.  As the specific distribution of the data in feature spaces has impact on efficient learnability of knowledge, the group studied this dependency in the realm of graphs~\cite{conf/ismis/DoerfelHS18} and in the realm of metric measure spaces~\cite{hanika2018intrinsic}.
  
The goal of this paper is to complements the existing, mostly conceptual work on CIL with some information on
\begin{itemize}
    \item the usage of terms such as information, knowledge, and experience as used in CIL (Section 2),
    \item remarks on scientific disciplines required in CIL research (Section 3), and
    \item a differentiation of CIL from other research areas and structured research programs (Section 4).
\end{itemize}

\section{Definition of some important terms}
\label{sec:definitions}

\subsection{Information and knowledge sources}

In the existing literature on CIL, the terms \textit{information source} and \textit{knowledge source} were used to summarize very different ways to gather the information or knowledge needed for learning in CIL based systems. Possible sources that have to be considered include
\begin{enumerate}
\item the \textit{own sensors} of a system that deliver measurements that are analyzed (e.g., to reason about the own behavior for success analysis as needed in reinforcement learning), 
\item \textit{other smart systems} that might have similar or complementary tasks (e.g., wearables such as smart watches or smartphones),
\item \textit{humans} that could be asked different kinds of questions (e.g., to label observations, to provide conclusions for rule premises, or to confirm knowledge),
\item the \textit{Internet} with its various data bases (e.g., collections of labeled images) or social media, and
\item \textit{simulation systems} that may be exploited to provide labeled data.
\end{enumerate}
In addition, these systems will keep temporarily unneeded but potentially useful information or knowledge in mind (i.e., their memory) which can be re-used in various ways, e.g., by transductive or transfer learning processes. 
In different situations different information and knowledge sources will be accessible or available. 
However, information and knowledge come at a cost. 
While access to the Internet is quite cheap (provided that Internet access is possible), for instance, involving humans can be rather costly (e.g., in terms of working hours) and should be restricted to rare, but important situations. This trade-off regarding the exploitation of various information  or knowledge sources has to be learned, too.
Our CIL systems have to combine various information or knowledge sources as well as various machine learning (ML) mechanisms and decide upon an appropriate choice of sources and learning mechanisms depending on availability, costs, quality, type of query, etc.

\subsection{Differences between information, knowledge, and experience}

Also, terms such as \textit{information}, \textit{knowledge}, and \textit{experience} have to be characterized more precisely. We adopt the meaning of these terms from the field of data mining~\cite{FKS11,FPS1996}. Data mining can be seen as a multi-step process 
where in a first step raw \textit{data} (e.g., collected with sensors) are pre-processed to condense application-specific \textit{information} in attributes or features. Then, \textit{knowledge} is extracted, e.g., by building classification or regression models based on rules. 
By analyzing this knowledge off-line it is possible to come to a deeper \textit{understanding} of its working principles, e.g., by (depending on the knowledge model) investigating how far premises of contradictory rules ``overlap'' or by determining the fraction of the input space which is ``covered'' by a certain rule.
In a given application, it is possible to gain \textit{experience} in using the knowledge, e.g., by determining how often a rule is applied or by stating how often it is applied successfully. 
Understanding and experience will both support the efficient and effective application of knowledge. 
Our CIL systems will address these levels as follows, for example:
\begin{enumerate}
\item The CIL system itself may collect sensor observations and extract relevant attribute values (data and information).
\item The operational behavior of the CIL system may be based on classifiers containing rule systems (knowledge).
\item Humans can be asked to provide labels for attribute vectors or conclusions for autonomously generated rule premises (information and knowledge) or they can be asked to perform an action that generates information or knowledge (e.g., by going to a certain place).
\item Assessing gained knowledge before and in action may lead to self-awareness of CIL systems regarding the usefulness of knowledge and knowledge deficiencies (understanding and experience).
\end{enumerate}
As understanding and experience can be seen as a kind of meta-knowledge, the term knowledge will be used when we refer to the levels beyond data and information.

CIL systems also have to consider that data, information, and knowledge are uncertain. Here, \textit{uncertain} is used as a generic term to address aspects such as unlikely, doubtful, implausible, unreliable, imprecise, inconsistent, or vague~\cite{MS97}.

\section{Remarks on required disciplines in CIL research}

CIL systems self-organize their \textit{learning} process autonomously in the sense that
\begin{itemize}
	\item they assess their own knowledge to decide when this knowledge is not sufficient to cope with new kinds of situations arising at run-time,
	\item they connect to new information or knowledge sources (e.g., other smart systems and/or humans) and know which kind of information or knowledge they can obtain from which source,
	\item they initiate interaction steps to inquire for information or knowledge that they have realized they miss to act optimally,
	\item they assess the quality of information or knowledge sources and the quality, usefulness, topicality, etc.\ of information and knowledge they gather, 
	\item they exploit various machine learning (ML) mechanisms to increase their own knowledge, e.g., collaborative learning, semi-supervised learning, transfer learning, reinforcement learning, and active learning, and
	\item they interact with humans in various ways and even support humans in their learning and cooperation processes.
\end{itemize}

Livelong learning (referring here to the lifetime of the CIL system) is required in open-ended environments, i.e., environments where
\begin{itemize}
	\item 
	the collective of participating entities (CIL systems, humans, etc.) might be very heterogeneous,
	\item 
	the number of entities might be large and open in the sense that new entities may enter the collective while others leave,
	\item 
	the environment in which the entities work might be time-dependent in the sense that it changes its properties over time (i.e., it is time-variant), or
	\item 
	the entities of the collective may have various tasks (including tasks that require intricate interaction schemes) that might change over time.
\end{itemize}
To ensure that our CIL systems do ``more good than harm'', we also have to develop new techniques to analyze and validate such systems (cf.~\cite{KZ09}), to detect emergent behavior, and to control their behavior in a way that guarantees certain properties that have to be defined for the various application fields.

The self-organized active gathering of information and knowledge can be seen as a first stage of cooperation between CIL systems that improve their knowledge to better master their individual tasks. This has an immediate impact on the usefulness of such systems for humans. In addition, these smart systems offer new chances to develop truly \textit{collaborating} systems and even new collaboration mechanisms for humans:
\begin{shortitemize}
	\item Collectives of CIL systems may master complex tasks that the individual entities cannot cope with.
  \item Humans may profit from the ``collective intelligence'' of all entities including any combination of other humans, CIL systems, the Internet, etc.
	\item Collaboration processes of humans may be actively supported by CIL systems being aware of the humans' respective needs and their knowledge.
	\end{shortitemize}
These opportunities require appropriate human-machine \textit{interaction} interfaces in order to exploit humans as an important and very particular source of information or knowledge, to (pro-)actively provide them with all the information or knowledge they might need (cf.~\cite{Sch15}), and to afford humans to inquire knowledge from such CIL systems. 

Following this line of thinking, next to scientific contributions in machine learning, CIL requires new scientific contributions from the fields of socio-technical system design, collaboration, and pedagogics to leverage complementary benefits of human and machine intelligence~\cite{TOL17,DLE+18}. For this purpose the area of CIL systems that shall improve human learning is particularly promising. 
\begin{itemize}
 \item It allows to specifically leverage on the potential of collaboration with various knowledge sources as well as context-dependent provision of knowledge for enhancing humans' higher-order-learning skills. 
  \item CIL Systems will have the potential to learn in a self-organized way by exchanging knowledge with other knowledge sources (i.e., intelligent systems and humans) in two different, but inseparable manners: On the one hand, AI provides suggestions to humans so that they are able to behave in a well-informed manner (i.e., AI is in the loop of human intelligence). On the other hand, humans train and tune machine learning models and evaluate their outputs (i.e., human intelligence is in the loop of AI). 
  \end{itemize}
The combination of both ways of learning in CIL systems and the resulting dependencies and connections of learning processes lead to the fact that understanding and designing CIL systems are highly complex endeavors. To increase CIL systems’ efficiency and maximize mutual benefits, it is important to enable learning effects in both directions by the means of human-machine collaboration. Therefore, learning processes of humans need to be facilitated, too. As a consequence, collaborative  procedures are required that refer to the work of two or more knowledge sources that work on common material towards a common group goal (e.g., human and non-human knowledge sources are learning as well). To achieve this goal, communication, coordination, and cooperation are required.

Altogether, we can state that various disciplines have to cooperate to pave the way for this new kind of smart systems.

	
\section{Differentiation from other on-going programs on related topics}
\label{sec:differentiation}

CIL is still a new research area. Substantial preliminary work, on which CIL will be based, has been accomplished in several research fields, some of which were already subject to in DFG (German research foundation) funding (e.g., priority programs Autonomous Learning and Organic Computing) or EU funding (e.g., research initiative Fundamentals of Collective Adaptive Systems). The following part initially summarizes related research areas and highlights open challenges that will be addressed by CIL. Afterwards, on-going programs are listed together with their relation to CIL.

\subsection{Related Research Areas}

\textit{Active Learning (AL)} provides powerful approaches to create flexible systems which are able to adapt themselves to a changing environment~\cite{Set12}.
These methods interact with their target system to investigate which information might best optimize their model, and they actively acquire this information. 
In classification (also in regression) problems, AL algorithms actively request the target value of an instance
\cite{AKG+14}. 
In literature, three basic AL paradigms are distinguished: 1) query synthesis (the query instance is generated), 2) pool-based AL (the query is an instance from a pool of unlabeled instances), and 3) stream-based AL (instances successively appear and the AL algorithm decides if the label should be acquired)~\cite{AKG+14}. 
The most popular method is uncertainty sampling~\cite{LG94}, although it solely exploits the model by acquiring labels from instances near the classifier's decision boundary~\cite{Set12}. 
More sophisticated methods extend this approach by adding exploratory components, density information, or class priors~\cite{cvbs:RS13:INS}. 
Other approaches use the technique of version space partitioning by building classifier ensembles~\cite{AKG+14}, for example.
The usefulness of an upcoming label is then determined from the classifier's disagreement~\cite{Set12}. 
Decision theoretic approaches, such as expected error reduction~\cite{RM01} or probabilistic active learning~\cite{KKL+16}, try to determine the real usefulness in terms of classification performance improvement.
A further aspect of AL (especially when applied to activity and behavior recognition problems) is the ability to efficiently generate training data, since labeling is an expensive task~\cite{SLS08}. Here, AL methods have been combined with traditional semi-supervised learning techniques, such as self-training and co-training~\cite{SLS08}. 
AL has also been applied to reduce labeling costs 
in health applications on mobile devices~\cite{LRE10}.

Open challenges in the field of AL that will be in the focus of the CIL initiative are:
\begin{shortitemize}
	\item CIL has to deal with {uncertainty regarding the quality of information} and knowledge sources (we still assume that there are no malicious or deceptive experts that cheat or attack the active learner).
	\item There are several or even {many knowledge sources} (e.g., humans, other smart systems) with {different degrees of quality} (e.g., certainty) who {collaborate} to provide the active learner with the required information.
	\item The {information sources} label not only samples but also other kinds of queries to {provide knowledge at a higher level} (e.g., by assigning a conclusion to a presented premise of a rule).
	\item CIL has to know which kind of information or knowledge they can obtain from {which source at which cost}.
	\item CIL has to develop techniques for AL systems, where experts will benefit from the active learner by {receiving feedback} in order to improve their own knowledge.
\end{shortitemize}

\textit{Autonomous Learning (ATL)} aims at increasing the degree of autonomy regarding machine learning in technical systems \cite{hammer2015special}. These systems collect data from their environment, autonomously choose parameters and representations, and interact with their environment. For that purpose, techniques such as reinforcement learning, unsupervised and semi-supervised learning, or active learning have to be adopted and advanced \cite{hammer2012challenges}. 
Open challenges in the field that will be in the focus of the CIL initiative are:
\begin{shortitemize}
	\item CIL has to go far beyond the objectives of Autonomous Learning as it requires livelong collaborative learning approaches for distributed, heterogeneous systems in open-ended environments.
	\item CIL does not focus on individual learning techniques -- instead it aims at exploiting different available techniques at runtime.
	\item Current activities in the field of ATL only consider -- if at all -- (individual) humans in Active Learning settings. In contrast, CIL will make use of crowds possibly as a mixture of humans and machines.
	\item Most of the activities in the field of ATL use robotic scenarios as basis for the investigations. In contrast, CIL aims at a generally applicable technological basis that is utilized in different application domains.
	\item CIL will go far beyond ATL since it will make use of {many knowledge sources} with {different degrees of quality and uncertainty} -- which is not in the scope of ATL, yet.
\end{shortitemize}

\textit{Collaborative Learning (CL)} is a field that aims at improving the learning behavior of individual systems by means of taking advantage of the knowledge of other systems (typically of the same kind) \cite{Hoen:LAMAS2005:MultiagentLearning}. This includes the exchange and fusion of knowledge, the transfer of knowledge, experiences, and observations, and the agreement on division of labor in learning tasks \cite{Panait:JAAMAS2005:MultiAgentLearnig}. In particular, concepts such as transfer learning and multi-agent learning are research directions in the field.
Open challenges in the field that will be in the focus of the CIL initiative are:
\begin{shortitemize}
	\item CL scenarios typically do not consider dynamics in the availability of knowledge sources or corresponding cost functions.
	\item CIL will go beyond CL in terms of handling unknown learning problems.
	\item Instead of transferring and customizing knowledge models in a transfer learning sense, CIL will develop novel cooperation schemes that, e.g., make use of crowds and their particular capabilities.
	\item In contrast to CL, CIL will explicitly model the uncertainties of the different cooperation partners and consider this information within the decision and learning processes.
\end{shortitemize}

Fundamentals of \textit{Collective Adaptive Systems (CAS)} investigate design and operation principles for heterogeneous, distributed systems with entities that have individual goals and solution strategies. These entities interact at various temporal and spatial levels \cite{choi2001supply}. A key aspect is the cooperation of humans with systems. Important issues are conflict resolution, long-term stability, handling noisy or outdated information, and development of open systems where single entities leave the overall system and new ones enter \cite{lansing2003complex}. Open challenges in the field that will be in the focus of the CIL initiative are:
\begin{shortitemize}
	\item Machine learning, in particular active and collaborative learning, are not in the focus.
	\item An individual assessment of uncertainties based on the experiences with the particular entity are not in the focus of CAS by now.
	\item CAS are consisting of autonomous entities and they interact with each other. However, these interactions have the purpose of problem-specific and task-oriented cooperation and typically do not include knowledge exchange and learning behavior.
	\item A major goal in CAS is the analysis of system behavior at macro-level resulting from interactions at micro-level. This has an emphasis on the sequence of system states and mostly neglects a goal-oriented self-improvement for lifelong learning capabilities.
\end{shortitemize}

\textit{Multi-Agent Systems (MAS)} is a field concerned with the design and cooperation schemes for distributed collections of autonomous subsystems \cite{wooldridge2009introduction}. In this context, an autonomous subsystem is called ``agent'' since it acts on behalf of a certain user and aims at achieving the predefined goals of this user. Conceptually, the term MAS summarizes research on questions in the context of how autonomous, distributed, and smart agents can share their knowledge and experience, negotiate their goals, and develop plans based on their potentially heterogeneous capabilities, resulting in coordinated actions and collective problem solving \cite{platon2007mechanisms}. Open challenges in the field that will be in the focus of the CIL initiative are:
\begin{shortitemize}
	\item The major idea of MAS is that agents acts on behalf of users. Consequently, the direct participation of humans in the system is neglected in most cases.
	\item Consequently, MAS does not include efforts on human-machine cooperation and crowd interfaces.
	\item If learning is considered in MAS these learning processes are typically clearly defined (e.g., using just reinforcement learning for a certain control problem), while CIL systems are expected to deal with high uncertainties regarding the learning problem. Furthermore, CIL techniques will allow for bilateral learning processes.
	\item MAS have an emphasis on the dynamics of the environment and the agent constellations located in these environments. However, the explicit modeling of uncertainties and the utilization of different agents as explicit source of knowledge do not play a major role.
\end{shortitemize}

\textit{Online or stream learning (OL)}~\cite{Lug11,Gam10} is a machine learning paradigm developed to work in time-variant (also called non-stationary or evolving) environments \cite{fontenla2013online}. 
Thereby, it delivers real-time predictions, efficiently built on large data streams (e.g., sensor inputs). 
One of the most important components is the ability to detect drift (abrupt change) or shift (gradual change). Here, change detection mechanisms are of great importance~\cite{EP11,GZB+14}. 
Online learning has also been used in smart home environments for activity recognition~\cite{KC+14} or to detect lighting behavior~\cite{DH14}. 
In the field of preference learning, the goal of a classifier is to predict the preferences of humans. 
Therefore, it is necessary to identify so-called perennial objects, e.g., in form of user profiles~\cite{STS+15}. 
Online algorithms have also been used in combination with AL to reduce labeling cost in a changing scenario~\cite{Lug17a,KKS15}.
Open challenges in the field that will be in the focus of the CIL initiative are:
\begin{shortitemize}
	\item OL does not influence the behavior of the learning source. In particular, it does not have control on the incoming data from the stream. In contrast, CIL actively selects and controls knowledge sources.
	\item OL does not incorporate system design and the corresponding processes of interactions.
	\item In contrast to OL, CIL will explicitly model the knowledge sources, their expertness in regions of the input space, and the reliability of their results.
	\item OL typically considers a static set of homogeneous data streams -- in contrast, CIL will focus on a dynamic selection and utilization of various heterogeneous knowledge sources.
\end{shortitemize}

\textit{Organic Computing (OC)} (see also Autonomic Computing (AC) \cite{KC03}) addresses complex technical systems that will self-adapt to new environmental conditions at runtime \cite{MST17}. Key technologies are self-$*$ techniques inspired by nature (e.g., self-configuration, self-organization, or self-optimization) \cite{schmeck2005organic}. Open challenges in the field that will be in the focus of the CIL initiative are:
\begin{shortitemize}
	\item OC lacks fundamental research on collaborative learning approaches (in particular active learning approaches) allowing for an adaptation to emerging environmental situations in time-variant environments as needed for CIL.
	\item OC and AC systems consider humans as sources for system goals and utility functions but do not consider them as explicit interaction partners in mutually beneficial processes.
	\item Typically, OC systems do not have the ability to assess their own knowledge.
	\item CIL will go far beyond OC since it allows for a proactive selection and exploitation of knowledge sources (while OC uses pre-defined knowledge sources in a static manner).
\end{shortitemize}

\textit{Collaboration Engineering (CE)} focuses on collaboration, the work of two or more individuals on common material, deliberately designed to achieve a common group goal~\cite{VBM09}. 
CE is a design methodology for the development of collaborative work practices to execute high-value recurring tasks, and deploying the designs of collaborative work practices to  practitioners to execute for themselves without ongoing support from professional facilitators~\cite{KV06}. 
The heart of the CE design methodologies is the Six-Layer-Model. To develop collaborative work practices, i.e., a recurring design of a collaborative process, it considers six different levels of abstraction (i.e., collaboration goal, group products, group activities, group procedures, collaboration tools, collaborative behavior). 
At each level, there are different phenomena of interest, and thus different design concerns, metrics, theories, modeling conventions, design patterns that need to be considered~\cite{BKV+14}. 
Consequently, CE constitutes an established design methodology with a broad field of research to design IT-supported collaboration among humans. 
However, current research challenges of CE refer to collaborative situations in which teammates are represented by humans and smart systems~\cite{SBB+18}. 
Open challenges in the field that will be in the focus of the CIL initiative are:
\begin{shortitemize}
	\item CE lacks guidelines for designing collaborative work practices among humans and smart systems. 
	Therefore, extensions of the CE methodology are needed, that will cope with situations in which a teammate is not human.
	\item CE needs exemplary designs and prototypes of collaborative work practices with teammates represented by humans or smart systems (such as CIL systems).
	\item Criteria need to be defined to decide whether a task should be executed by humans or machines or hybrid forms. This will lay a foundation for the applicability of using CE methodologies.
\end{shortitemize}

\textit{Crowdsourcing} describes the outsourcing of tasks to an independent mass of people by an open call via the Internet on IT-facilitated platforms~\cite{DBL16,How08,EG12}. 
A specific form of crowdsourcing is crowd work, which constitutes a digital form of gainful employment by which the crowd workers receive a monetary payment~\cite{BMB+18,DBL16,KNB+13}. 
This allows to access and mobilize human expertise and competences~\cite{SG11} 
and can be used to perform information or knowledge related tasks with low costs~\cite{SG11}. 
Typical tasks in the context of crowd work are labeling images, translating or transcribing text, and providing opinions or ideas. Since crowd work can be applied within and beyond the organizational boundaries, one can distinguish between internal and external forms. In internal settings, the own employees act as an internal crowd to elevate the human chapter of the company. In contrast, external forms of crowd work include individuals from outside the company. Existing IT-facilitated platforms such as Amazon Mechanical Turk (MTurk) offer opportunities for an open call and easy distribution of tasks~\cite{IPW10}. 
However, in such settings, the task design constitutes a critical success factor. Open challenges in the field that will be in the focus of the CIL initiative are:

\begin{shortitemize}
	\item The design of tasks in crowdsourcing initiatives to support active learning.
	\item Solutions and reference processes for improving crowdsourcing activities by using CIL technologies.
	\item CIL technologies will go beyond existing selection strategies and will develop selection strategies to identify crowd workers with the appropriate expertise in order to minimize costs (e.g., for labeling a task). 
\end{shortitemize}

\textit{Socio-technical systems} are systems that involve a complex interaction between humans, machines, and the environmental aspects of a work system. Nowadays, this interaction is true of most enterprise systems~\cite{ET60}.  
For developing such systems, socio-technical systems design (STSD) methods are used to consider aspects with regard to people, machines, and context. The development of such socio-technical systems, therefore, is subject to major challenges. For these reasons, it is of great importance that system developers receive guidelines and design patterns that support them in developing socio-technical systems to take into account all factors. Open challenges in the field that will be in the focus of the CIL initiative are:

\begin{shortitemize}
	\item In a broader sense CIL systems constitute socio-technical systems. In this light, CIL will define socio-technical design requirements for CIL systems. 
	\item CIL will define socio-technical design patterns to build CIL systems.
\end{shortitemize}

\subsection{Research programs}

CIL will be based on results of a number of already completed or running research programs. These programs addressed either methodical aspects, foundations, or potential application fields:

\subsubsection*{DFG Programs:}

This section addresses programs of the German Research Foundation (DFG). The abbreviations used in the following are CRC (Collaborative Research Centre, German: SFB), PP (Priority Program, German: SPP), RTG (Research Training Group, German: GRK), TRR (Transregio), and RU (Research Unit, German: FOR). 

The \textit{PP 1183} (\textit{Organic Computing}) investigated various self-adaptation and self-organization mechanisms for the design and operation of complex technical systems. This also included novel techniques for (cooperative) learning. However, the projects did not explicitly consider humans as participants in the system structure and neglected challenges resulting from assessing knowledge or interacting with crowds.

In the \textit{PP 1593} (\textit{Design for Future -- Managed Software Evolution}), the focus is set on long-living systems and consequently related to the vision of CIL where life-long learning capabilities are investigated. The main idea in the PP 1593 is to make legacy systems ``stay young'', i.e., make them able to self-adapt to changes in requirements or environment, respectively. However, CIL technology is closing a gap in the broader scope of the PP since collaborative learning and interaction with humans (or crowds) has been neglected so far.

The \textit{PP 1527} (\textit{Autonomous Learning}) focuses on machine learning techniques that work widely independently from human experts (see also previous section about ``Related Research Areas''). The goal is to reach autonomy regarding parametrization, learning objectives, choice of knowledge representations, selection of knowledge sources, etc. 

The \textit{PP 2037} (\textit{Scalable Data Management for Future Hardware}) investigates novel data management solutions that allow for more sophisticated data manipulation methods and closer interweaving between operating system and data management system. Although having a clearly different scope, possible input may become relevant for CIL in terms of providing different levels of abstraction for data with the vision to hide certain functionalities depending on the context. However, the PP just started and concrete results on this topic are still open.

The \textit{PP 1999} (\textit{Robust Argumentation Machines, RATIO}) provides a forum for research on novel methods for extracting arguments and relationships as well as semantic models and ontologies from documents. This also includes search algorithms and representation techniques. Although analyzing data and developing models, the scope of the PP differs largely from CIL as neither collaboration nor interaction play a role.

The \textit{PP 1835} (\textit{Cooperative Interactive Cars}) aims at improving the interaction between vehicles. This involves car-to-x communication and supports autonomous driving. In the context of CIL, results from this PP may become relevant for explicitly investigating cooperation schemes and selecting cooperation partners. However, the concepts of learning and interaction with, e.g., humans plays a minor role.

Quite closely related to the PP 1835 is the \textit{RTG 1931} (\textit{SocialCars: Cooperative (De-)centralised Traffic Management}) that tackles challenges in the area of vehicular traffic management and incorporates approaches for collecting relevant spatial information, providing communication infrastructures, and including human and economic influences into the system.

The projects within the \textit{PP 1324} (\textit{Extraction of Quantifiable Information from Complex Systems}) focused on mathematical methods for extracting quantifiable information from large complex systems, e.g., from physics. The projects neither take the distributed nature of systems we consider here into account nor do they explicitly consider humans as participants in the system structure.

The \textit{PP 1736} (\textit{Algorithms for Big Data}) combines the perspectives of recent hardware developments (i.e., technological challenges to appropriately capture better computational models) and problem specific algorithms (i.e., challenges due to massive data availability). Although making use of machine learning technology and being closely related to CIL in terms of the underlying possibly large amount of data, the scope is clearly different as CIL neglects questions related to hardware or processing challenges.

The \textit{RU 1085} (\textit{Trustworthiness of Organic Computing Systems}) investigated mechanism for robustness of open, heterogeneous collections of autonomous systems. The projects developed novel techniques for determining trustworthiness of these systems, for selection of interaction partners based on computational trust, and for normative control of autonomous systems. The developed concepts provide valuable insights for aspects of the selection process for interaction partners in the context of CIL.

The \textit{RU 1800} (\textit{Controlling Concurrent Change}) addresses a question that is of some importance to CIL: How can independent software updates be made in increasingly open and interconnected systems while different applications share resources? The goal is to manage updates without side effects and to guarantee the functionality of software subsystems. CIL systems may deal with common knowledge models that can be considered as such shared resource. 

The \textit{CRC/TTR 62} (\textit{Companion Technology}) works on methods for building cognitive companion systems supporting individual users. The main focus is on the interaction between human users and their companion systems, but does not go beyond this local scope.

The \textit{RU 1513} (\textit{Hybrid Reasoning for Intelligent Systems}) aims at integrating qualitative and quantitative forms of reasoning, which is envisioned to result in hybrid reasoning formalisms. The RU is related in terms of developing intelligent systems -- but focuses on reasoning instead of learning.

The \textit{RU 2535} (\textit{Anticipating Human Behavior}) focuses on capabilities of computer systems to classify image or video clips taken from the Internet or to analyze human pose in real-time for gaming applications. The vision is to develop novel methods that proactively anticipate the behavior of humans who interact with the computer to allow for minimal delays in the interaction behavior. 

The \textit{TRR 248} (\textit{Basic of understandable software systems -- for a comprehensible cyber-physical world}) aims at developing novel technology for understanding how complex systems interact with each other and which decisions actually caused the observed behavior. The vision is to provide basics for root cause analysis and automatic explanation of decisions.


Authors of this paper (Lukowicz, Zweig, Sick) have played a key role in several of these programs, e.g., Organic Computing, Autonomous Learning, Cooperative Interactive Cars, 
or Algorithms for Big Data.

\subsubsection*{BMBF Programs:} This section focuses on programs funded by the German Federal Ministry of Education and Research (BMBF).

The \textit{``Gr\"undungen: Innovative Start-ups f\"ur Mensch-Technik-Interaktion''}
program funds start-up projects developing digital solutions for digital
societies. The here proposed \emph{BoInHo2020} project develops an intelligent
bot supporting the students in university teaching realm. Research conducted in the field of 
CIL may contribute largely to such projects, which are not unlike,
but essentially different, to demonstrators that are addressed by CIL.

The goal of the recent \textit{``Richtlinie zur F\"orderung von Zuwendungen f\"ur
	Forschung zur Gestaltung von Bildungsprozessen unter den Bedingungen des
	digitalen Wandels (Digitalisierung II)''} is the comparison of
teaching-related study outcomes and degrees, as well as equal access and equal
recruitment in teaching practice and profession. This involves the development of digital
systems combining theoretical and practical teaching.

\textit{``Richtlinie zur Förderung von Forschung zu ethischen, rechtlichen und
	sozialen Aspekten (ELSA) der Digitalisierung, von Big Data und Künstlicher
	Intelligenz in der Gesundheitsforschung und –Versorgung''} is another recent
call. Projects funded by this initiative shall work on methods for developing scientifically and
technologically well-founded frameworks for analyzing and evaluating health and
care related research in the era of big data. A particular challenge here is the
interaction and cooperation between humans and AI driven machines, e.g., in the
form of decision support systems, to which CIL methods are essential.

The \textit{``Richtlinie zur Förderung von Forschungsvorhaben im Rahmen der
	Innova\-tions- und Technikanalyse''} includes multiple topics related to CIL.
Those are \emph{Digital Work Design} and \emph{Akzeptanz humanoider
	Service-Roboter}. They investigate collaborative working schemes as well as
characterizations of acceptable human-machine interaction.

The \textit{``Richtlinie zur Förderung von Projekten zum Thema
	’Weiterentwicklung der Indikatorik für Forschung und Innovation’ ''} aims at
improving the instruments of innovation measurement, to develop it further and
to adapt it to dynamics in the field of innovation. This includes every part of
the innovation process, even human capital. Here, CIL technology may enable so
far not used collaborative approaches for measurement.

The \textit{``Richtlinie im Rahmen der Strategie der Bundesregierung zur
	Internationalisierung von Bildung, Wissenschaft und Forschung zur Förderung
	von Vorhaben der strategischen Projektförderung mit der Republik Korea unter
	der Beteiligung von Wirtschaft und Wissenschaft im Bereich Robotik''} focuses
on knowledge sharing and joint development of robotics with the Republic of
Korea. The funded projects include industrial and scientific research on robots
interacting with humans. This includes interactive learning schemes as developed
in the field of CIL.

The \textit{``Richtlinien zur Förderung von Maßnahmen für den
	Forschungsschwerpunkt ’Zukunft der Arbeit: Arbeit in hybriden
	Wertschöpfungssystemen’ im Rahmen des FuE-Programms ’Zukunft der Arbeit’ als
	Teil des Dachprogramms ’Innovationen für die Produktion, Dienstleistung und
	Arbeit von morgen’ ''} combines the perspective of technical and social
innovation in companies and organizations. The CIL related goal here is the
development of socially relevant hybrid methods.

The \textit{``Digitale Medien in der beruflichen Bildung (Fachprogramm)''}
investigates methods to support the use of digital media, Web 2.0 technologies,
and mobile applications in the context of state-approved training occupation and
education. Methods for collaborative interactive learning are highly demanded in
these settings.

\subsubsection*{BMWI Programs:}

The following section focuses on programs funded by the German Federal Ministry for Economic Affairs and Energy (BMWI).

The goal in \textit{``PAiCE - Platforms/Additive
	Manufacturing/Imaging/ Communication/Engineering''} is to study the transfer of
present digital technologies from the lab to the market, such as
engineering collaboratively across companies in the project
INTEGRATE. Research results from the field of CIL may contribute to such research
projects by providing new transferable lab technologies.

The \textit{``Smart Data – Data Innovations''} program includes projects that
employ crowdsourcing for mobility solutions. However, those projects rather focus on
data enriching than collaborative learning.

The \textit{``Smart Services World – Internet-based Business Services'' (I/II)} program
promotes projects that build prototype solutions based on networked smart
technical systems. Research in this program is conducted only in the fields of legal
questions, standardization, and the security of platform architectures.

Projects in \textit{``Autonomics for Industry 4.0''} are mainly industry
related. They do implement (known) human-machine interaction methods based on
expertise and therefore profit from novel methods developed by CIL.

\subsubsection*{EU Programs:}

The following section focuses on programs of the European Union (EU).

The vision of distributed, intelligent, interacting systems being part of everyday living environments has a long research tradition in the EU Future and Emerging Technologies (FET) Unit. 
It goes back to the \textit{Disappearing Computing} initiative\footnote{\url{http://www.disappearing-computer.org}} which had its final report in 2007. Its stated  mission has been ``to see how information technology can be diffused into everyday objects and settings, and to see how this can lead to new ways of supporting and enhancing people's lifes'', which has been the precursor to much of today's ubiquitous computing and Internet of Things. One of the authors (Schmidt) had been part of this initiative pioneering some key concepts of context aware interactive systems \cite{schmidt2000implicit}. 

Following a few years later, the COSI-ICT (\textit{Science of Complex Systems for Socially Intelligent ICT}) call of 2009 attempted to go beyond the impact on individuals towards the study of interaction between networked intelligent systems and people on social level combining computer science and complex systems approaches. One of the authors (Lukowicz) coordinated  the SOCIONICAL project within this initiative where the core work on collaborative indoor location as an initial example of large scale ad-hoc collaboration and model fusion emerged~\cite{kloch2011emergent}.

Focusing more closely on collaboration between human and digital ensembles was the  recent FET FOCAS (\textit{Fundamentals of Collective Adaptive Systems}\footnote{\url{http://www.focas.eu}} initiative. One of the authors (Lukowicz) was part of the \textit{ALLOW ENSEMBLES}\footnote{\url{https://cordis.europa.eu/project/rcn/106345/factsheet/en}} and the \textit{Smart Society}\footnote{\url{https://cordis.europa.eu/project/rcn/106959/factsheet/en}} projects within this call. The work resulted in initial collaborative learning model fusion approaches focusing on high level fusion. 

As a follow up in 2018, the  FET proactive \textit{Socially Interactive Technologies} has started with the aim of developing ``socially interactive media with radical improvement for building trust and understanding, social integration, engagement, collaboration, learning, creativity, entertainment, education and wellbeing, among others''.
Another related initiative is the \textit{HumanE AI} preparatory action for a new FET Flagship that is coordinated by two of the authors (Lukowicz and Schmidt). The project aims at a new generation of ethical AI technologies that can enhance human capabilities and has goals similar to CIL. However, it focuses on the development of a research agenda and long term vision rather than concrete research. 

In summary, the work on CIL is rooted in some previous  and running initiatives of the EU in which authors of this paper have played a key role. However, while related on a high level of abstraction and general goals, the specific direct research goals of all those initiatives were different from CIL.

\bibliographystyle{splncs04}
\bibliography{references}

\end{document}